\newcommand{\CLASSINPUTtoptextmargin}{20mm}
\documentclass[letterpaper, 10 pt, conference, twocolumn]{ieeeconf}

\IEEEoverridecommandlockouts
\usepackage{cite}
\usepackage{amsmath,amssymb,amsfonts}
\usepackage{multirow}
\usepackage{algorithm}
\usepackage{algorithmic}
\usepackage{graphicx}
\usepackage{textcomp}

\usepackage{multirow}
\usepackage[table,xcdraw]{xcolor}
\usepackage{soul}
\usepackage{subcaption}

\def\BibTeX{{\rm B\kern-.05em{\sc i\kern-.025em b}\kern-.08em T\kern-.1667em\lower.7ex\hbox{E}\kern-.125emX}} % [RHT] Adding some color definitions
\definecolor{deep-red}{RGB}{192, 0, 0}
\definecolor{deep-purple}{RGB}{120, 0, 170}
\definecolor{good-green}{RGB}{0,135,0} 
\definecolor{purple}{RGB}{210, 0, 210}

\newcommand{\ch}[1]{\textcolor{black}{#1}}

\makeatletter
\let\NAT@parse\undefined
\makeatother
\usepackage[colorlinks]{hyperref}

\begin{document}

% Adjust spacing around floats (reduce it)
\setlength{\textfloatsep}{8pt plus 2pt minus 4pt}

\title{Realistic Data Generation for 6D Pose Estimation of Surgical Instruments} \author{%
     Juan Antonio Barragan$^{1}$, Jintan Zhang$^{1}$, Haoying Zhou$^{2}$,\\ Adnan Munawar$^1$, and Peter Kazanzides$^1$
     \thanks{$^1$Department of Computer Science, Johns Hopkins University, Baltimore, MD 21218, USA.  
     Email: {\tt jbarrag3@jhu.edu, jzhan247@jhu.edu}}
     \thanks{$^2$Department of Robotics Engineering, Worcester Polytechnic Institute, Worcester, MA 01608, USA.  
     Email: {\tt hzhou6@wpi.edu}}
}

\maketitle
 \begin{abstract}
Automation in surgical robotics has the potential to improve patient safety and surgical efficiency, but it is difficult to achieve due to the need for robust perception algorithms. In particular, 6D pose estimation of surgical instruments is critical to enable the automatic execution of surgical maneuvers based on visual feedback. \ch{In recent years, supervised deep learning algorithms have shown increasingly better performance at 6D pose estimation tasks; yet, their success depends on the availability of large amounts of annotated data. In household and industrial settings, synthetic data, generated with 3D computer graphics software, has been shown as an alternative to minimize annotation costs of 6D pose datasets. However, this strategy does not translate well to surgical domains as commercial graphics software have limited tools to generate images depicting realistic instrument-tissue interactions.} To address these limitations, we propose an improved simulation environment for surgical robotics that enables the automatic generation of large and diverse datasets for 6D pose estimation of surgical instruments. Among the improvements, we developed an automated data generation pipeline and an improved surgical scene. To show the applicability of our system, we generated a dataset of 7.5k images with pose annotations of a surgical needle that was used to evaluate a state-of-the-art pose estimation network. The trained model obtained a mean translational error of 2.59\,mm on a challenging dataset that presented varying levels of occlusion. These results highlight our pipeline's success in training and evaluating novel vision algorithms for surgical robotics applications. 
\end{abstract}

\section{Introduction}
In minimally invasive robotic surgery, automation of time-consuming and repetitive surgical subtasks has the potential to reduce the surgeon's mental demands and improve the overall efficiency of surgery \cite{attanasioAutonomySurgicalRobotics2021}. Automation of surgical subtasks has been extensively studied by the research community, leading to autonomous algorithms for suturing \cite{schwanerAutonomousNeedleManipulation2021, wilcoxLearningLocalizeGrasp2022a, jiangMarkerlessSutureNeedle2023,chiuMarkerless2022}, blood suction \cite{richterAutonomousRoboticSuction2021, barraganSACHETSSemiAutonomousCognitive2021a}, and tissue retraction \cite{luSuPerDeepSurgical2021}, among others. \ch{One key challenge of surgical automation is developing perception algorithms to compensate for the robot’s kinematic inaccuracies and execution failures. This requires estimating the 6D pose of rigid and articulated instruments from endoscopic video to modify autonomous motions based on visual feedback. }  

\ch{In the task of 6D pose estimation, the goal is to estimate the translation and rotation of the object of interest with respect to the camera coordinate frame}. This task has traditionally been approached by extracting 2D visual features from RGB images and then matching them with corresponding 3D features on the object's model. \ch{These 2d-3d correspondences can then be used as an input to a Perspective-n-Point \cite{Fischlerrandom1981} solver to retrieve the object's pose.} 

%and Ransac algorithms, the pose can be retrieved. 
% This information can be later used to grasp the object with the robot manipulator or provide a visual overlay

More recently, end-to-end deep neural networks have demonstrated superior performance for 6D object pose estimation tasks than traditional point-pair feature approaches \cite{sundermeyerBOP2023}. The main drawback of these deep learning approaches is the need to generate large amounts of annotated training data, which, for 6D pose estimation tasks, is prohibitively expensive to obtain. As a solution, it has been shown that high-fidelity synthetic data of models of physical objects can be used to train pose networks that perform well in locating their real counterparts \cite{sundermeyerBOP2023, tobinDomain2017}. 

In the surgical context, synthetic data generation is a more challenging endeavor as it is important to generate samples that portray sensible instrument motions and realistic tissue-instrument interaction. Currently available simulation environments such as Vision Blender \cite{cartuchoVisionBlender2020} or BlenderProc \cite{denningerBlenderProc}  can be utilized to render annotated images of surgical instruments in surgical backgrounds; however, they offer limited capabilities on how the objects can be moved or interact with each other in the scene. Furthermore, they do not offer good support to work with articulated robotic instruments.  

%% SOMETHING TO ADD LATER
% Lastly, to translate algorithms trained in simulation to reality, it is important that the virtual environment is made up of models of physical objects that can easily be acquired by researchers, which is often not the case for off-the-shelf simulation environments.

%Lastly, to promote reproducible results in autonomous surgical robotics research, it is important that experimental setups can be easily reproduced. In this regard, synthetic data generation environments should be based on physical objects that can easily acquired by researchers, which is often not the case for off-the-shelf simulation environments.

% Furthermore, simulated environments should be based on physical objects that can be easily acquired by surgical robotics researchers, such that networks trained with the synthetic data can later be easily translated to physical setups. All these constraints currently limit the use of state-of-the-art neural networks for the tasks 6-DOF detection of surgical instruments.

 In previous work, an open-source platform for surgical suturing was introduced to address some of the limitations of surgical robotics simulation platforms \cite{munawarOpen2022}. In particular, this work, built with the Asynchronous Multi-Body Framework (AMBF)\cite{munawarambf2019}, introduced improved robot control algorithms and teleoperation capabilities, and provided access to ground-truth imaging data. Although it enabled the collection of realistic suturing motions via teleoperation, it still lacked the capabilities to automatically generate the large-scale datasets needed for neural network training. Furthermore, the scene provided a simplified phantom, not resembling any real physical phantom, which complicated the task of physically reproducing the virtual environment.
 
 To address these limitations, we have developed an automated data generation pipeline on top of \cite{munawarOpen2022} to produce large-scale and diverse datasets from pre-recorded trajectories generated with teleoperation. Moreover, we scanned and added a commercially available training suturing pad to our simulation environment to improve realism and to ensure that the virtual scene could be physically reproduced. The goal of these improvements was to facilitate the creation of the large-scale datasets needed for deep learning-based algorithms.    
 
 \begin{figure*}[ht]
      \centering
      \includegraphics[width=1.0\textwidth]{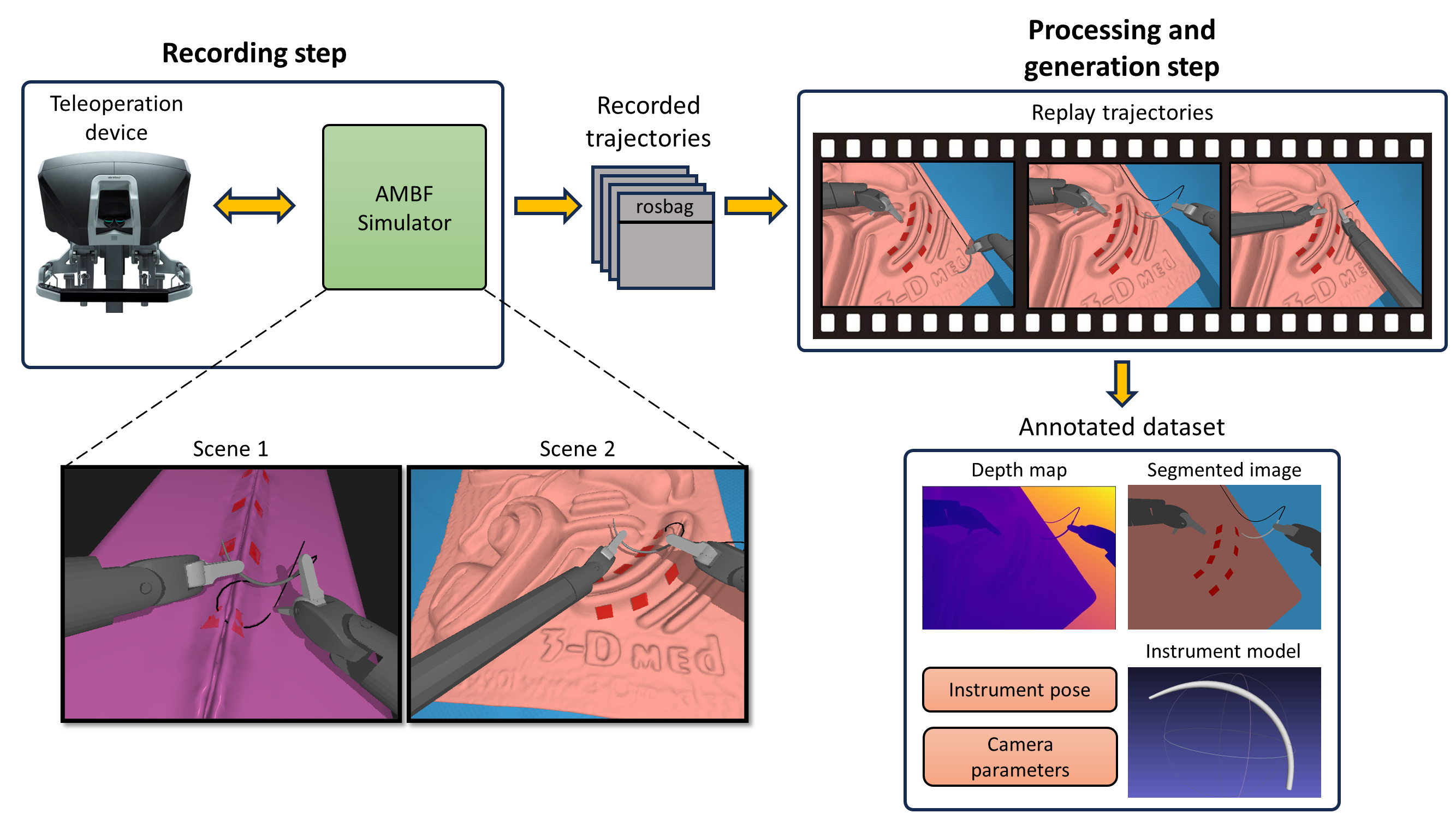}
      \captionof{figure}{Proposed data collection pipeline. Realistic data generation with our proposed system requires two steps. First, trajectories of the robotic manipulators are collected using a teleoperation device. Second, trajectories are replayed automatically multiple times from different camera viewpoints to generate diverse set of images. While replaying, our pipeline stores depth and segmentation maps and the ground truth pose of all the objects with respect to the camera.}
      \label{fig:system_overview}
 \end{figure*}
 
\ch{In this paper, we showcase an application of our pipeline by generating data to train a state-of-the-art 6D pose estimation network to predict the pose of a needle while performing suturing maneuvers.} Regarding the network architecture, the GDR-Network \cite{wangGDRNet2021} was chosen as it is one of the first fully differentiable networks for the task of pose estimation from monocular RGB. We envision that our work will significantly benefit the community of surgical robotics researchers as we provide a standardized platform for generating, evaluating, and deploying novel vision algorithms. Lastly, we highlight that thanks to the modular nature of the base simulation engine, our data generation pipeline can be used for a wide range of tasks and objects in surgical robotics setups. 

In summary, this work presents the following contributions:
\begin{enumerate}
    \item An automated data generation pipeline for 6D pose estimation of surgical instruments.
    \item A realistic simulation environment for surgical suturing based on a commercially available suturing pad model.
    \item A dataset of 7.5k images with 6D pose annotations for a simulated surgical needle.
    \item  \ch{Evaluations on a state-of-the-art 6D pose estimation neural network on the task of surgical needle pose estimation.}
\end{enumerate}
 
\section{Related Work}

\subsection{Offline Simulation Data Generation}
% -- Vision Blender
Vision Blender \cite{cartuchoVisionBlender2020} is a Blender add-on for generating synthetic computer vision data such as RGB, depth, segmentation, optical flow, and surface normals. Designed as a tool to efficiently generate data for surgical robotic system development, Vision Blender supports converting generated data to Robot Operating System (ROS) \cite{Quigley2009ROSAO} messages. %messages so that developers can evaluate the algorithms with the synthetic data as if they are collected from the physical robotic system.  With its graphical user interface, Vision Blender is friendly to individuals with limited programming experience.
% -- BlenderProc
Another modular procedural pipeline for generating simulation data based on Blender is BlenderProc \cite{denningerBlenderProc}. It shares data generation capabilities with Vision Blender, with the added feature of bounding box generation.  BlenderProc also supports importing data in URDF format, expanding its capability for modeling complex robotics systems. Compared to using the native Blender Python, which demands a deep understanding of the Blender infrastructure,  BlenderProc offers an intuitive Python interface to simplify the scene-building and data-acquisition process.
% -- UnrealROX
% In addition to the tools developed based on Blender, Martinez et al.\cite{Martinez2019unrealrox} presented an approach based on Unreal Engine 4 that renders photorealistic scenes and robots into a virtual reality headset and stores per-frame scene information for offline data and ground-truth generation.

\subsection{Robot Simulator \& Digital Twins}
% -- Isaac Sim
% [What is Isaac Sim]
NVIDIA® Isaac Sim is a scalable robotics simulator and synthetic data generator. 
% [What is its feature?]
Powered by the GPU-accelerated physics simulation engine PhysX and physically-based rendering technology Iray, 
Isaac Sim is capable of simulating physically accurate virtual environments and generating photorealistic data. 
Isaac Sim offers support for Universal Scene Descriptor (USD) and Unified Robot Description Format (URDF), 
enabling developers to seamlessly import intricate 3D environment definitions and robot configurations with ease.
In addition, Isaac Sim allows developers to establish connections between Isaac Sim and their custom robot applications 
via integrated Robot Operating System (ROS1 \& ROS2) interfaces. 
With extensions such as Isaac Gym \cite{makoviychuk2021isaac} and Isaac Orbit \cite{mittal2023orbit}, 
developers can efficiently test and refine their robotic systems and robot learning algorithms.
% [What works have been done Isaac Sim]
% For example, Marinho et al. \cite{marinho2022design} created an AI robotic platform prototype, its digital twin in Isaac Sim, 
% and performed validation experiments such as peg transfer, gauze cutting, and drilling for a cranial window under teleoperation. 

% -- AMBF
Defining multi-body robots using formats like URDF or Standard Description Format (SDF) can lead to ambiguous definitions in cases of densely connected, sparsely connected, or unconnected bodies.
% [What is AMBF]
To address this constraint, Munawar et al.\cite{munawarambf2019} introduced the Asynchronous Multi-Body Framework (AMBF), an innovative front-end description format for multi-body simulation, aimed at simulating complex closed-loop robots. 
% [How it does simulation?]
AMBF leverages Bullet Physics \cite{Bullet_Physics_Simulation} for its physics simulation and CHAI-3D \cite{MUSIC202010216} for graphics rendering and haptic volume rendering. Within AMBF, every object features a custom OpenGL shader, facilitating diverse data generation capabilities, including RGB, depth, and segmentation maps.

% [What works are done based on AMBF?]
Building upon AMBF, many research efforts have emerged to advance robot-assisted surgeries. These include using AMBF to design image-guided feedback modalities\cite{ishida2023improving}, a novel framework for skull base surgeries featuring high-precision optical tracking and real-time simulation \cite{ShuDigital2023}, and a causality-driven robot tool segmentation algorithm \cite{cartsding}. % [What are we doing with AMBF] 
In this work, we selected AMBF simulation over other simulation alternatives due to its support of a broad array of input devices, which facilitates the collection of realistic motions of the surgical instruments. In particular, its tight integration with the da Vinci Research Kit (dVRK) \cite{kazanzidesOpensource2014} allows collecting robotic surgical motions with a similar setup to what is used in surgery.

\section{Methodology}

The primary motivation of this work was to provide a data generation tool for 6D pose estimation of surgical instruments. In this regard, we adapted an open-source simulation environment to automatically generate sequences of images of robotic-assisted surgical actions with their corresponding ground-truth maps. The proposed data generation pipeline 
 (See figure \ref{fig:system_overview}) was developed with the goal of generating programmatically large and diverse datasets. The methodology section is divided as follows. In section \ref{subsect-data_collection}, we present the pipeline for automatic data generation. Section \ref{subsect-virtual_scene} shows the improvements in the surgical virtual scene. Section \ref{subsect-dataset} describes the generation of a dataset for the task of needle pose estimation. Section \ref{subsect-deep_models} describes the pose estimation deep learning model trained to estimated the needle's pose. Lastly, section \ref{subsect-evaluation metrics}, describes the evaluation metrics used for the predictions of the trained neural network.

\subsection{Data generation pipeline} \label{subsect-data_collection}

Our proposed data generation pipeline is composed of two stages: a recording step, and a processing and generation step. During the data recording step, a teleoperation device is used to move the virtual robotic manipulators to perform the surgical task. While teleoperating, joint and Cartesian positions of the robotic manipulators, and the poses of other objects in the simulation are stored in a rosbag file\footnote{A rosbag is a file format used to store messages, from the Robot Operating System (ROS) middleware. It is ideal for storing trajectories from a robot.}. 
    
During the processing and generation step, the stored robotic trajectories are replayed multiple times under different camera viewpoints and lighting conditions. While replaying the trajectories, a collection script stores the resulting monocular or stereoscopic RGB images with their corresponding ground-truth information, i.e., depth map, segmented images, camera intrinsic parameters, and pose of objects expressed with respect to the camera coordinate frame.         

\subsubsection{Format for generated data}

To store the data, it was decided to use the Benchmark for 6D Object Pose Estimation (BOP) format \cite{sundermeyerBOP2023}. This is a standardized format adopted by several benchmark 6D pose estimation datasets such as HOPE \cite{tyree6DoF2022}, YCB \cite{xiangPoseCNN2018}, and others. Moreover, it is a standard format used for an annual 6D pose competition \cite{sundermeyerBOP2023}. In the BOP format, related data are grouped under a \textit{scene\_id}. For our pipeline, data from each trajectory replay was stored in a different \textit{scene\_id}.

\begin{figure}[ht]
    \centering
    \includegraphics[width=0.48\textwidth]{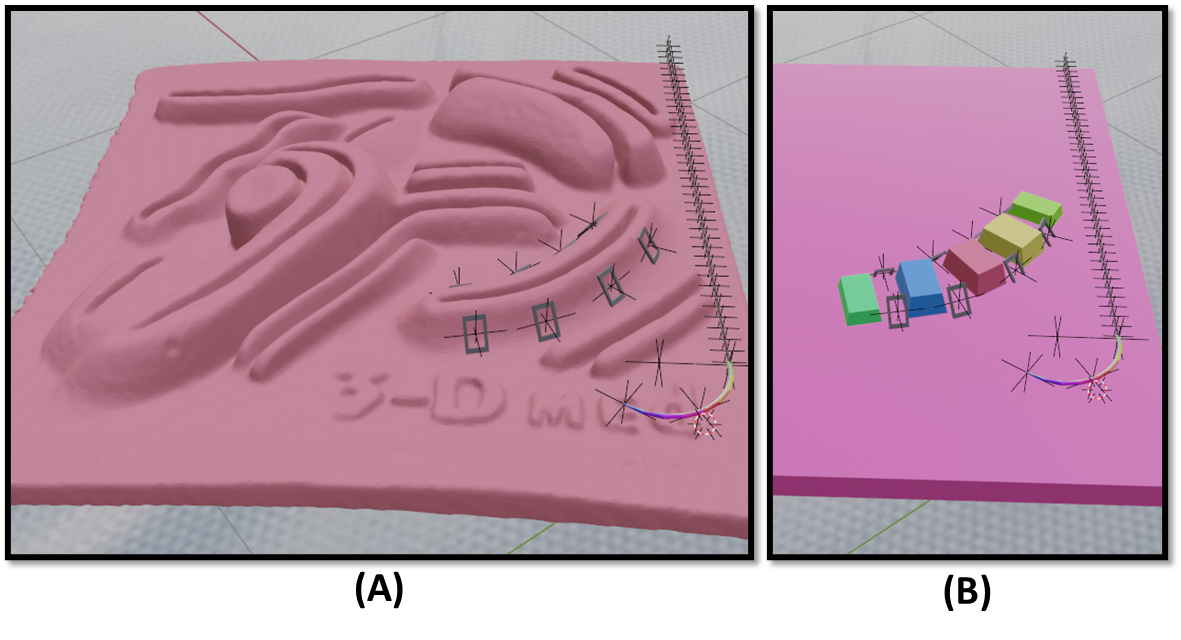}
    \caption{(a) Visual mesh of the 3-Dmed phantom after preprocessing. (b) Simplified collision mesh composed of multiple convex subcomponents assembled into a single mesh. The collision mesh was only provided for a single ridge of the phantom.}
    \label{fig:3-Dmed_phantom}
\end{figure}

\subsection{Improvements of virtual scene} \label{subsect-virtual_scene}

To improve the realism of our virtual scenes, a commercially available suturing pad (3-Dmed, Franklin, OH, US) was added to the simulation. The suturing pad was initially MRI scanned to obtain a mesh that was preprocessed using 3D Slicer \cite{Fedorov_3d_slicer_2012} and Meshlab \cite{vittorio_meshlab}. The MRI scanning was selected over other modalities as it provides higher contrast for soft tissue phantoms\cite{hiorns2011imaging}. Using the resulting mesh, an AMBF Description File (ADF) is made by utilizing the Blender-AMBF addon plugin \cite{munawarOpen2022}. As observed in figure \ref{fig:3-Dmed_phantom}, the full-resolution suturing pad is used for visualization, while a simplified mesh made of convex subshapes is used for collision. Small corridors are left on the collision mesh to allow for needle insertions similar to the scene developed in \cite{munawarOpen2022}. Collision meshes are simplified to optimize the simulation's performance. 
% ADDITIONAL PROCESSING STEPS
% \textcolor{blue}{**Jack's note: \begin{enumerate}
%     \item why MRI instead of CT? $\rightarrow$ Compared to CT scanning, MRI scanning can provide more details and have higher contrast for soft tissues or phantoms\cite{hiorns2011imaging}
%     \item processing step of the mesh : 1. obtain the raw DICOM file from the MRI scanning [the output is consisted of 82 slices(along z direction, total height 32.4mm) and each slice has a shape of $\mathbb{R}^{512 \times 512}$ and the side length of the grid is 0.3516mm] 2.reconstruct the volume based on 3D slicer pre-defined segmentation method and manually modify the ROI to filter out the outliner 3. export the result as STL/OBJ files from 3D slicer. 4. import the exported file to meshlab and then clearn up \& repair the vertices and faces.
% \end{enumerate}}

\begin{figure}[ht]
    \centering
    \includegraphics[width=0.45\textwidth]{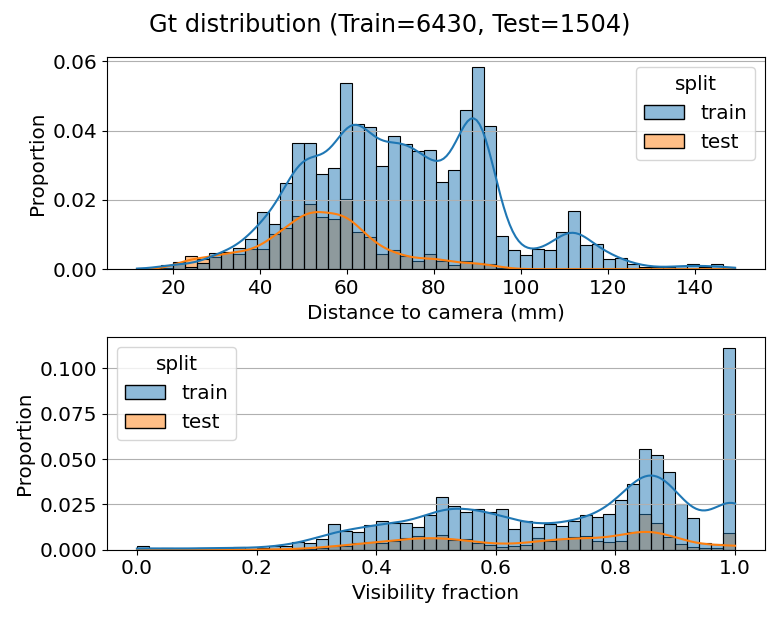}
    \caption{Ground-truth distribution of the collected needle 6DoF detection dataset.}
    \label{fig:gt_distribution}
\end{figure}

\begin{figure*}[ht]
      \centering
      \includegraphics[width=0.85\textwidth]{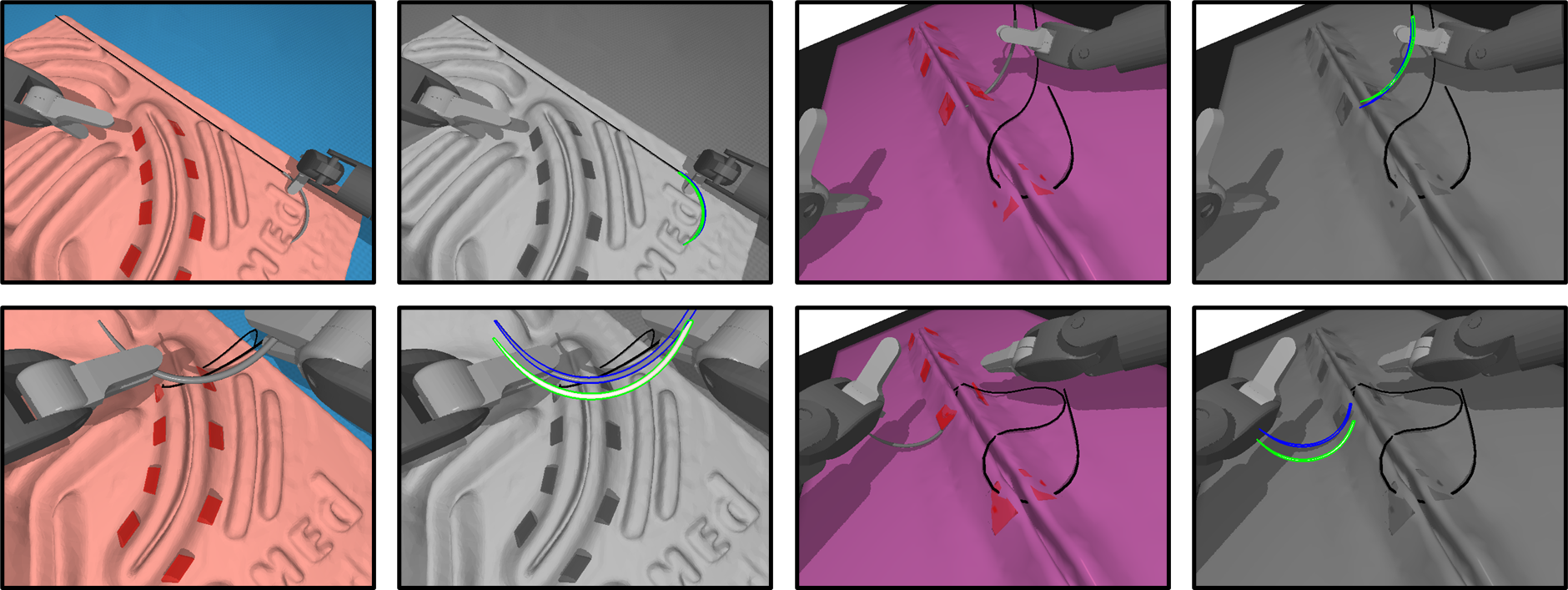}
      \captionof{figure}{Test set sample frames and corresponding pose prediction visualizations. Colored images show samples from the test dataset. Masks in the grayscale images are generated by projecting the needle model to the image with the ground-truth (blue mask) and the network's estimated pose (green mask). Higher overlaps between the green and blue masks are indicative of better pose estimates.}
      \label{fig:dataset_examples}
 \end{figure*}

\subsection{Generated dataset for surgical needles pose detection} \label{subsect-dataset}

Using our improved simulation environment, we collected a dataset for the task of 6D pose estimation of an 18.65\,mm surgical needle. First, we collected 6 rosbag recordings of suturing motions using a dVRK robot's surgical console \cite{kazanzidesOpensource2014}. Two recordings were done in scene 1 and four in scene 2 (See figure \ref{fig:system_overview}). 

Each of the 6 collected recordings was then replayed on the simulator 20 times, each time from different camera positions and view angles. Data from 4 recordings were used as a training set and 2 for the testing set. Camera positions were specified in the joint space of a virtual endoscopic camera manipulator (ECM) provided by the base simulation environment. To produce unique viewpoints with every replayed recording, a small random offset was added to the selected ECM joints. 

After filtering images where the needle was not present, 6430 training and 1500 testing images with a 640x480 resolution were obtained. As observed in figure \ref{fig:gt_distribution}, the resulting dataset is more challenging and realistic than the one presented in \cite{jiangMarkerlessSutureNeedle2023} as the needles in the images present varying levels of occlusion and distance to the camera. The visibility fraction is calculated with

\begin{equation}
   \text{visibility} = \frac{\text{area of visible mask}}{\text{area of projected mask}}
\end{equation}

\noindent
\ch{where the visible mask is the set of pixels in the RGB image that correspond to the needle and the projected mask is the set of pixels obtained by projecting the needle's CAD model to the image plane using the ground-truth pose.}
 
\subsection{Selected deep learning model for 6D pose estimation} \label{subsect-deep_models}

Using the dataset described in section \ref{subsect-dataset}, we trained the state-of-the-art network for 6D pose estimation GDR-Net \cite{wangGDRNet2021}. This network was selected as it is one of the first fully differentiable pose estimation methods in the literature and the winner of the BOP pose estimation competition of 2022 \cite{sundermeyerBOP2023}. This network receives as an input a 2D RGB region, where the object of interest is located, and outputs three intermediate geometric feature maps: the \textit{visible object mask}, a map of 2d-3d dense correspondences, and a surface region attention map. These intermediate maps are concatenated and then given as input to a fully-differentiable Patch-PnP module that regresses the final rotation and translation of the object. 

Training of this network can be performed in an end-to-end manner and only requires the RGB image and the object CAD model to generate ground truth for the intermediate geometric maps. As mentioned above, the network requires a region of interest (ROI) where the object is located. To obtain these ROIs during our experiments, the off-the-shelf detector YOLOX \cite{Ge2021YOLOXEY} was also trained with our dataset for the task of 2D  bounding box detection for the needle. 

\begin{figure}[ht]
    \centering
    \includegraphics[width=0.45\textwidth]{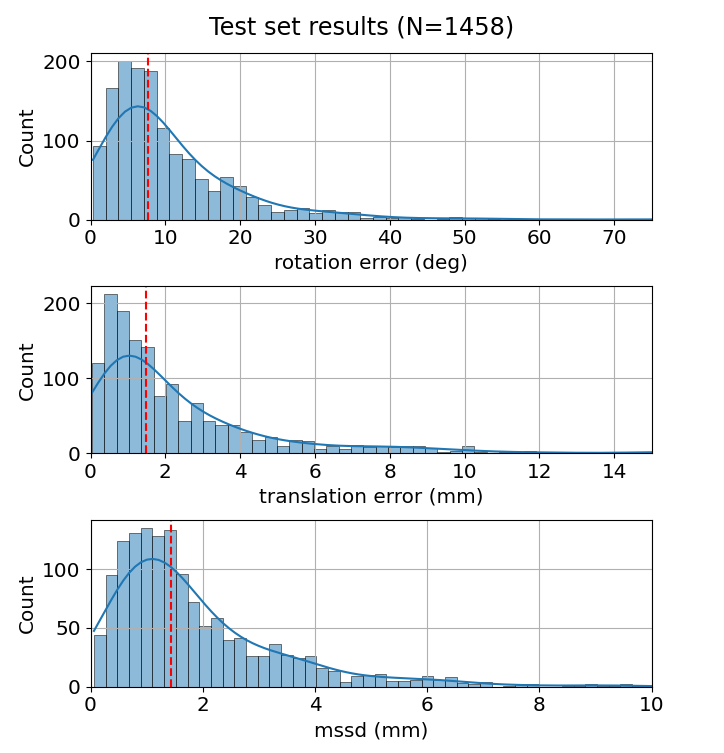}
    \caption{Error distribution for the best GDRNet model on the test dataset. The x-axis represents the different error metrics, and the y-axis is the number of samples within each bin. The red dotted line indicates the median performance. The x-axes of the histograms were truncated respectively at 70\,mm, 15\,deg and 10\,mm for visualization purposes.}
    \label{fig:pose_error_distrib}
\end{figure}
 
\subsection{Evaluation metrics for the pose estimates}\label{subsect-evaluation metrics}

Pose estimations from the neural network were evaluated using three common error metrics: (1) translation error ($e_{TE}$), (2) rotation error ($e_{RE}$)\cite{Hodan2016OnEO}, and (3) Maximum Symmetry-Aware Surface Distance ($e_{MSSD}$) \cite{Drost_Introducing_2017}. \ch{Metric 1 measures the translational error using the Euclidean distance. Metric 2 measures the rotational error using the axis angle representation of rotation matrices. Lastly, metric 3 measures the maximum distance between a vertex of the object model transformed with the ground truth and estimated pose.}  Given a ground truth pose $\bar{P}=(\bar{R},\bar{t}$), an estimated pose $\hat{P}=(\hat{R},\hat{t})$, and a set of vertices $V_M$ belonging to the object model, the metrics $e_{TE}$, $e_{RE}$ and $e_{MSSD}$ can be calculated with 

\begin{equation} \label{error_pos}
 e_{TE} = ||\bar{t} - \hat{t}||   
\end{equation}

\begin{equation} \label{error_rot}
 e_{RE} = arccos((Tr(\bar{R}\hat{R}^{-1}-1)/2)   
\end{equation}

\begin{equation} \label{error_MSSD}
 e_{MSSD} = \min _{\mathbf{S} \in S_M} \max _{\mathbf{x} \in V_M}\|\hat{\mathbf{P}} \mathbf{x}-\overline{\mathbf{P}} \mathbf{S x}\|_2 
\end{equation}
\noindent
where $S_M$ is a set of symmetry transformations for the object whose pose is being estimated.

\begin{figure*}[ht]
    \centering
    \includegraphics[width=0.90\textwidth]{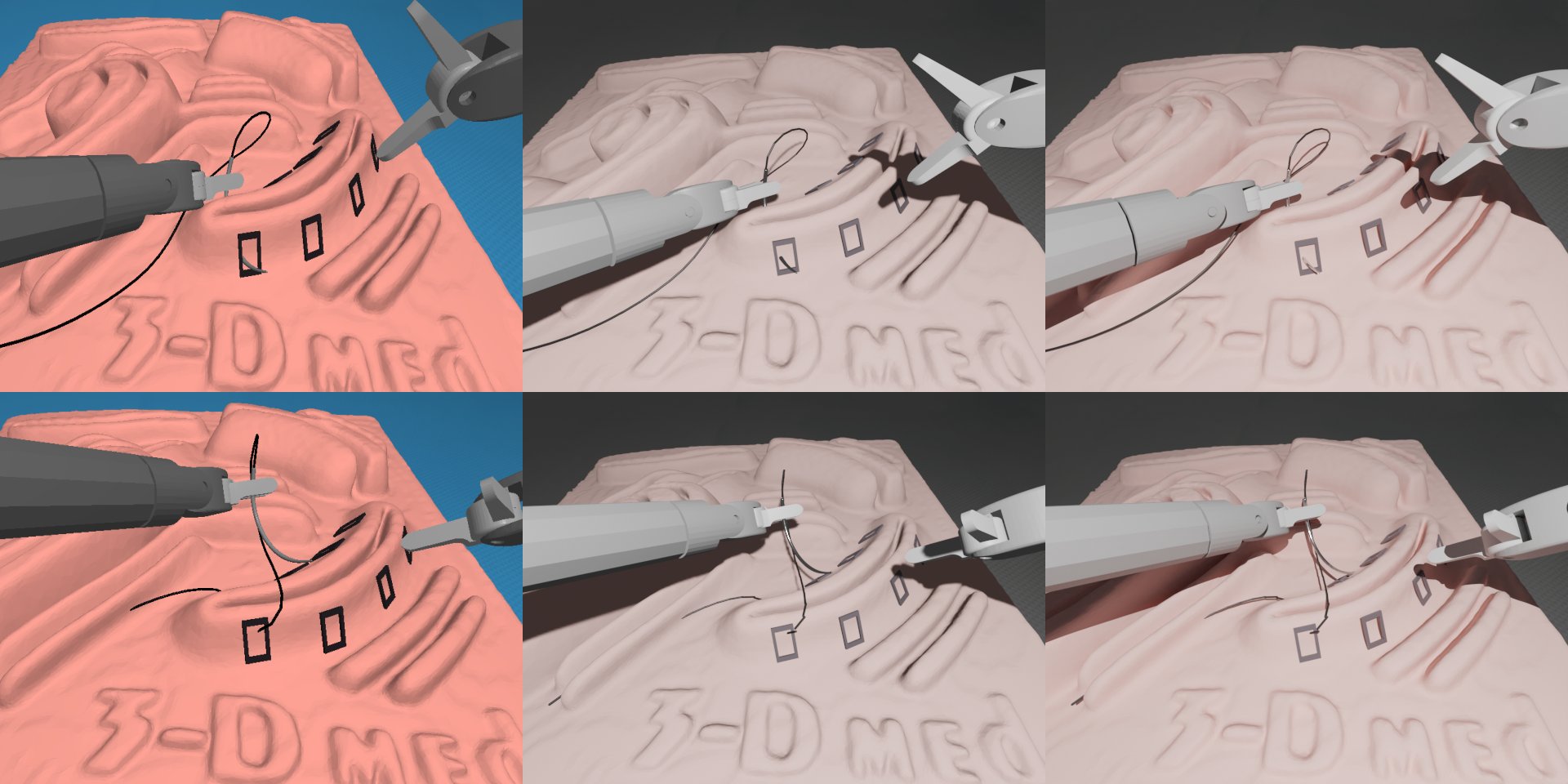}
    \caption{Rendering quality comparison between AMBF (left), Eevee (center), and Cycles (right). Each row represents the same scene. Shadow quality and metal shininess are superior in Cycles due to more comprehensive and exhaustive ray tracing, while the needle is less glossy and the shadow is unrealistically uniform in Eevee. Nevertheless, both Eevee and Cycles produce significantly higher fidelity rendering than AMBF.}
    \label{fig:blender_render_needle_inside}
\end{figure*}

\section{Experiments and Results}

For the evaluation experiments, first, the YOLOX and GDR-Net networks were trained with the generated training dataset. YOLOX was trained for 30 epochs using the Ranger Optimizer\cite{RangerOpt}, a batch size of 16 and a learning rate of 1e-3. GDR-Net was trained with the Ranger Optimizer for 450 epochs, a batch size of 48 images and a learning rate of 8e-4. This training setup was similar to the one used in \cite{wangGDRNet2021}. At test time, the trained bounding box detector was used to predict a single region of interest for each image. This region of interest was then used as input for the GDR-Net. Only images where at least 30 percent of the needle was visible were used for evaluation. Some sample images from the test set with their corresponding pose predictions can be observed in figure \ref{fig:dataset_examples}.

Final pose errors can be seen in table \ref{tab:gdrn_results_table}. On the evaluated images, GDR-Net obtained a median rotational error of 7.74 degrees and a median translation error of 1.49\,mm (less than 20 percent of the needle's diameter). These results are comparable to the pose detection results of non-occluded needles presented in \cite{jiangMarkerlessSutureNeedle2023} even though needles in our test set present varying levels of occlusion. Lastly, the median MSSD error is 1.43\,mm. Pose error distribution in figure \ref{fig:pose_error_distrib} indicates that the network can have sporadic predictions with significantly higher errors. \ch{High pose errors can be mainly attributed to images where several needle poses cannot be distinguished from each other.}

\begin{table}[ht]
\centering
\begin{tabular}{|cccc|}
\hline
\multicolumn{4}{|c|}{\textbf{Test set results (N=1458)}}                                                                                                                                \\ \hline 
\multicolumn{1}{|c|}{\textbf{}}       & \multicolumn{1}{c|}{\textbf{$\mathbf{e_{RE}}$ (deg)}} & \multicolumn{1}{c|}{\textbf{$\mathbf{e_{TE}}$ (mm)}} & \textbf{$\mathbf{e_{MSSD}}$ (mm)} \\ \hline
\multicolumn{1}{|c|}{\textbf{mean}}   & \multicolumn{1}{r|}{11.85}                            & \multicolumn{1}{r|}{2.59}                           & 2.09                              \\
\multicolumn{1}{|c|}{\textbf{std}}    & \multicolumn{1}{r|}{17.52}                            & \multicolumn{1}{r|}{3.41}                           & 2.43                              \\
\multicolumn{1}{|c|}{\textbf{median}} & \multicolumn{1}{r|}{7.74}                             & \multicolumn{1}{r|}{1.49}                           & 1.43                              \\
\multicolumn{1}{|c|}{\textbf{min}}    & \multicolumn{1}{r|}{0.33}                             & \multicolumn{1}{r|}{0.04}                           & 0.06                              \\
\multicolumn{1}{|c|}{\textbf{max}}    & \multicolumn{1}{r|}{170.5}                            & \multicolumn{1}{r|}{33.01}                          & 20.91                             \\ \hline
\end{tabular}
\caption{GDRNET test set results. Only images where at least 30 percent of the needle was visible were included in the evaluation. The diameter of the detected needle was 18.65\,mm.}
\label{tab:gdrn_results_table}
\end{table}

%%%%%%%%%%%%%%%% 
% the needle has a radius of 10.18mm, therefore we can claim that the needle has a diameter of 20mm/21mm if no further scaling is implemented. I believe 20mm/21mm is still acceptable for laparoscopy surgery

\section{Discussion and Future Work}

In this work, we developed a data generation pipeline for 6D estimation tasks of surgical instruments on top of the simulation framework AMBF. The proposed pipeline generates monocular or stereoscopic RGB images, and pose annotations for any rigid or articulated instrument in the scene. Moreover, each generated RGB image is accompanied by its corresponding depth and segmentation maps.

The focus of the work was to enable the automatic generation of large and diverse datasets showing realistic tissue-instrument interaction and sensible trajectories for robotic manipulators. In this regard, we divide our data generation pipeline into two steps: (1) a data recording step where robotic trajectories of a surgical task are recorded, and (2) a processing and generation step where each collected trajectory is replayed multiple times from different camera view angles and lighting conditions. 

To showcase the applicability of our pipeline, we generated a dataset of 7.5k images with pose annotations for a surgical needle to evaluate a state-of-the-art pose estimation neural network. After training, the network had translation and rotation errors comparable to previous works \cite{jiangMarkerlessSutureNeedle2023,chiuMarkerless2022} while being tested on a challenging dataset where the needle could be partially occluded by the instruments and the tissue. 

\ch{Although the network showed good performance on average, it is important to remember that the model makes predictions solely based on the visual appearance of the object, and therefore cases where multiple poses are indistinguishable from each other will result in high pose errors. Specifically for surgical needles, there are two main scenarios leading to pose ambiguities: (1) images where both the needle’s tail and tip are occluded and (2) images where the needle’s curvature cannot be observed, i.e., the needle appears as a straight line. As a solution, pose ambiguities could be resolved by using the network’s predictions with a model-based tracker that uses additional priors, such as the robot's kinematic motion or the pose of the needle in previous frames.}

In future work, we will leverage our data generation pipeline to study different techniques for transferring pose detection models from simulation to reality, a problem that is often referred to in the literature as the ``domain gap'' \cite{schramlPhysically2019}.  As noted by \cite{sundermeyerBOP2023}  and \cite{herediaperezEffects2020}, rendering realism plays an important role in transferring neural networks from synthetic to real objects. This hints that models trained based on our current synthetic data (generated with the simpler Blinn-Phong shading technique \cite{BlinnModels1977} ) might suffer from degraded performance when applied to data from the physical surgical platform. 

To mitigate this limitation, we implemented a preliminary real-time pipeline to improve the rendering quality by transferring the object pose (including cameras and lights) from the AMBF simulator to Blender. This allows us to utilize the two state-of-the-art rendering engines included in Blender since version 3.0: Eevee (a real-time rasterization-based renderer) and Cycles (a physically based path tracer).  As shown in figure \ref{fig:blender_render_needle_inside}, shadows and metal shininess rendered using Blender are significantly better than AMBF. Future studies will focus on understanding the effects of different rendering algorithms on the simulation-to-real transfer of neural networks. Additional future improvements on our simulation platform will include more accurate models for the robotic instruments and advanced materials that more accurately reflect surgical tools.

\section*{Acknowledgments}
This work was supported in part by NSF AccelNet awards OISE-1927354 and OISE-1927275. We thank Irene Kim, Haochen Wei, and Nicholas Greene for their invaluable insights on 6D pose estimation models. 

\section*{Supplementary information} 
For more information, visit the project repository at \href{https://github.com/surgical-robotics-ai/realistic-6dof-data-generation}{https://github.com/surgical-robotics-ai/realistic-6dof-data-generation}

\bibliography{ICRA2024-SRC}
\bibliographystyle{IEEEtran}

\end{document}